# Morphological study of Albanian words, and processing with NooJ

Odile Piton[1], Klara Lagji[2]

[1] University Paris1 Panthéon-Sorbonne  [2] University of Tirana

## Abstract

We are developing electronic dictionaries and transducers for the automatic processing of the Albanian Language. We will analyze the words inside a linear segment of text. We will also study the relationship between units of sense and units of form. The composition of words takes different forms in Albanian. We have found that morphemes are frequently concatenated or simply juxtaposed or contracted. The inflected grammar of NooJ allows constructing the dictionaries of flexed forms (declensions or conjugations). The diversity of word structures requires tools to identify words created by simple concatenation, or to treat contractions. The morphological tools of NooJ allow us to create grammatical tools to represent and treat these phenomena. But certain problems exceed the morphological analysis and must be represented by syntactical grammars.

## Introduction

The late reform, in 1972, has lead to the current Albanian Literary Language that is the result of the "unification" of two Albanian dialects: Gheg and Tosk. This work focuses on the building of Natural Language Processing tools that are useful for the processing of the Albanian language. The dictionaries needed by automatic treatment must register basic vocabulary. They must be associated with tools that are able to recognize words obtained according to creative paradigms. So these tools "compute" the new words of the constructed part of the dictionary. Our purpose is to be able to recognize words automatically in a text and some morphological features. A general feature of Albanian is that there are a lot of "words with particles in two units" that will be very troublesome to tag. They are not compound words, and you cannot delay the work on this as a second step, after a first step that would be just recognizing simple or compound words. This is true for part of verbal forms, for most adjectives and for some pronouns and nouns. Another point is that, like every language, Albanian has some creative paradigms, from derivation rules to concatenation rules that are very active and *produce concatenated words, lacking in any dictionary.*

In the first section, we describe some distinctive features of the Albanian language. Then we describe how we use inflectional graphs, morphological graphs and syntactical graphs for the building of electronic dictionaries for the Albanian language. So this work focuses on morphological and syntactic features. We have managed to register the dictionaries in different formats. It is kept in tables, from which we generate lists of words for NooJ dictionaries. However, our tables can evolve and be adapted to the necessities of semantic tagging.

## 1. First Remarks on Albanian Language

Let us make a comparison between some of the most frequent tokens into an English book, a French book and an Albanian book. We display the number of words and the number of tokens[1] in Table 1.1.

**Table 1.1** Number of words and number of tokens in three books and three languages.

| Title | Author | Number of words | Tokens | Language |
|---|---|---|---|---|
| The portrait of a lady | Henry James | 229,000 | 11,000 | English |
| La femme de trente ans | Honoré de Balzac | 70,000 | 9,150 | French |
| Kush e solli Doruntinën[2] | Ismail Kadare | 14,980 | 8,738 | Albanian |

We have distinguished the five most frequent tokens. In the English book which are 13.6 per cent of the tokens and 13.3 per cent in the French book. See Table 1.2. We observe that they are determiners, propositions and conjunctions.

**Table 1.2.** Most frequent tokens into an English book and a French book.

| English text | | French text | |
|---|---|---|---|
| tokens | frequency | tokens | frequency |
| the | 3,25% | de | 4,08% |
| to | 3,13% | la | 2,96% |
| of | 2,66% | le | 2,31% |
| a | 2,31% | et | 2,14% |

---

[1] A token is a sequence of letters, it can be a word, or a contraction like *cannot*, or a component of an compound like *fur* in au fur et à mesure .

[2] " Who Brought Back Doruntine?" published in English as "Doruntine".

|   |       | I | 2,26% | les | 1,83% |
|---|-------|---|-------|-----|-------|
|   |       |   | 13,61% |     | 13,32% |

- **Most Frequent Tokens in the Albanian text:**

In the Albanian book, the three most frequent tokens are 13% of the whole. The three words are very frequent particles that are *part of articulated words or articulated adjectives.* See Table 1.3.

.**Table 1.3.** Most frequent tokens into an Albanian book.

| token | frequency | Category |
|-------|-----------|----------|
| të | 5,94% | Adjective and noun particle / verb particle / pronoun |
| e | 5,79% | Adjective and noun particle / pronoun/ conjunction |
| i | 2,21% | Adjective and noun particle / pronoun |
| total | 12,94% | |

Our purpose is not to display grammar of the Albanian language here, but it is important to list some precise points that have consequences on our objective: building dictionaries and transducers for the automatic treatment of the Albanian texts.

- **Albanian alphabet and alphabetical order**

The Albanian alphabet has 36 letters: 7 vowels and 29 consonants. It uses the Latin alphabet. The whole list is *a b c ç d dh e ë f g gj h i j k l ll m n nj o p q r rr s sh t th u v x xh y z zh*. This list is in alphabetical order. Nine of the consonants are written with digraphs[3] or double characters: *dh gj ll nj rr sh th xh zh*, they must be considered as one letter. But programs are not used to taking digraphs into account.

Collation, or assembly of written information into alphabetical order, is not similar to sorting algorithms: e and ë are not considered as consecutive letters. As for digraphs, it is different as well. Any word beginning with dh must be positioned after any word beginning with a d . However, when you alphabetize a list of words, any word beginning with di will be placed after dh . It is very time consuming to process big files with words that do not follow printed dictionaries order. This process requires extra time for each item that needs to be verified.

---

[3] Called digrams in Albanian, but digram means reoccurring two words in NooJ s environment.

## 2. Electronic Dictionary From Printed Dictionary

- **Printed Dictionary**

We have used the Albanian-French dictionary (4951 words) of the book Parlons Albanais [Ch Gut A Brunet-Gut Remzi Përnaska]. We have observed its format. Part of speech, such as noun, verb, etc. is not always written, but the Albanian speaker is generally able to know if an entry is a verb, a noun, an adjective, or an adverb  We transform the dictionary to prepare the dispatching of words into categorical lists.

**Table 2.1.** Format of the entries in the dictionary

| Category | Format | Examples |
|---|---|---|
| Noun | rad/t1,-t2 g, nb. (-t3, -t4) | aeroplan,-i m. pl. (-ë, -ët) plane m. |
| Active Verb | form1 (form2, past participle) | laj (lava, larë) to wash |
| Non active Verb | form1 (u form2, participle) | lahem (u lava, larë) to wash oneself |
| Adjective | form (i,e)<br>form(e) | mirë (i,e)  good<br>absurd(e) absurd |
| Preposition | entry POS  plus features | afër adv. and prep. + abl. near |
| Adverbs | entry adv. | pjesërisht adv. partly |
| Interjection | entry! excl! | adio! excl. adieu! |

We have used Excel with VBA macros to extract with few errors, 95% of the words of the list and to share out the different parts of the speech, on separated sheets (Fig. 2.1.). Lots of errors have been found, from minor ones, like a lack of a comma, space or hyphen, to some bigger ones. Some of these errors are never noticed by the reader, who is able to correct them mentally and automatically. He is able to supply slight deficiencies, but computers cannot! The best way, for an editor, to check a dictionary seems to be by using a program.

- **Electronic Dictionary**

We use Excel plus Macros to dispatch verbs, masculine nouns, feminine nouns, pronouns, adjectives, adverbs, prepositions and conjunctions on different sheets. Unrecognized items are inserted on a  problem  page that has to be processed manually. Then specialized Macro constructs entries for NooJ, according to the category and required features of each subset. NooJ s dictionaries accept single words as well as locutions or compound words with a hyphen or an apostrophe e.g.: projekt-ligj. Some words that include an apostrophe come from Turkish: Et hem, Mit hat.

**Fig. 2.1.** Distribution of words between different lists.

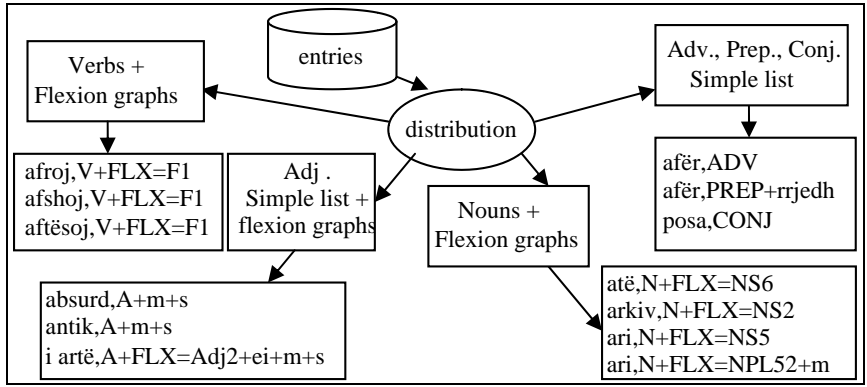

- **Prepositions, Adverbs, Conjunctions**

Conjunctions, adverbs and prepositions are invariant.

| | | |
|---|---|---|
| *absolutisht,ADV* | *së afërmi,ADV* | *porsa,CONJ* |
| *afër,ADV* | *afërsisht,ADV* | *posa,CONJ* |
| *afërmendsh,ADV* | *afro,ADV* | *posa që,CONJ +UNAMB* |

Some prepositions determine a case noted as a feature: kallez for accusative (kallezore), gjin for genitive (gjinore), rrjedh for ablative (rrjedhore) and emer for nominative (emërore).

*me anë,PREP+gjin*     *afër,PREP+rrjedh*     *tek,PREP+emer*

## 3. Building Dictionary with Inflectional graphs

### 3.1. Inflectional Graphs for Declension

NooJ has two tools to describe the flexions of a word: one is flexion files that "describe" the flexion; an other is flexion graphs that "draw" the flexion in a graph. We use graphs. Each declension is called 'FLX= Xn', where Xn is a declension graph with sub-graphs. Flexion description gives *radical + termination + flexion* for defined and undefined paradigms. The **Table 3.2.** shows an example.

## 3.2. Verbal system

In the dictionary, verbs are listed in the first person singular present tense. This form is called a lemma. A verb can have one or two forms: active form and/or not active forms: e.g. laj *to wash* and lahem *to wash oneself*. The two forms have the same past participle and four tenses of non-active forms are built on the same tense as the active form preceded by the particle "u": e.g. lava *I washed* (aorist) u lava *I washed myself*. There are twenty-one tenses. Some tenses use particles: "të" and "do".

In most cases, if X is the list of the preterit forms, "*të X*" gives the list of forms for a subjunctive tense and "*do të X*" gives the list of the forms for a conditional mode. It is similar between the subjunctive present tense that has the form "*të Y*" and simple future tense that has the form "*do të Y*", with the same list of forms Y. The Y list is very similar to the present tense, for verbs of the active form and it is exactly the same phenomenon for non-active verbs. We must add that clitics can be inserted before X or Y and after të. The declension of verbs is mainly regular, from three forms: present, aorist, past participle. They are written in the dictionary. Some tenses use particles 'të' or 'do të'.

We have drawn *inflectional graphs* to build one form for each person of each conjugation. That is to say 8 verbal forms, plus two forms for imperative. Active and non active verbs are described by separate graphs. The total is about 200 graphs and sub-graphs. We present the graph for the present tense of verbs like afroj. It is processed from left to right. <B> means delete the last character. The features written under the nodes are features that are added to the entry. See Fig. 3.1. The category "V" is for verb: afroj,V+FLX=Xi. Where "Xi" is the main graph and "PR_afroj" is a sub-graph for the present of the indicative.

**Fig. 3.1.** Graph for the present tense of verbs similar to afroj *to approach*.

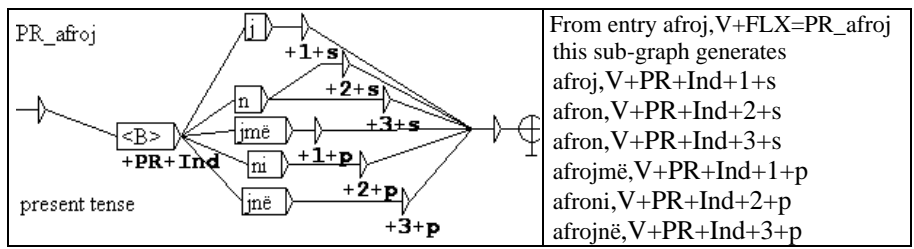

From entry afroj,V+FLX=PR_afroj this sub-graph generates
afroj,V+PR+Ind+1+s
afron,V+PR+Ind+2+s
afron,V+PR+Ind+3+s
afrojmë,V+PR+Ind+1+p
afroni,V+PR+Ind+2+p
afrojnë,V+PR+Ind+3+p

Meta-language commands are between brackets <>. The possibilities are to either delete a letter, add a letter, and go left or right inside the word. Insertions are in the nodes. The graph treats the input words, it deletes, or adds letters and adds features (information written under the nodes) separated by the + sign. +PR+Ind+3+p means present of indicative, third person of plural. The conjugation of afroj is the most regular. It applies to 30% of the verbs

As seen before, we have to take care of digraphs. For some verbs, aorist is marked by a change in vowel, e.g. e →o . We separate the set of these specific verbs into two subsets and we draw two graphs, according to the occurrence of one simple or double letter. See regular paradigm 4 in Table 3.2.

## 3.3. Inflectional Graphs for Nouns, Pronouns and Adjectives

- **Nominal and pronominal System (Words without Article)**

Albanian is an inflected language. Nouns are masculine, feminine or neutral. (Neutral nouns are rare, and they have no plural. We do not discuss them in this paper). Other nouns have four forms: two for the singular and two for the plural. One is called indefinite while the other is called definite. In the printed dictionary, these four forms are written in the nominative case: *rad/t1,-t2 g, nb. (-t3, -t4)*. E.g.: *an/ë,-a f. pl. (-ë, -ët) side*.

Masculine and feminine nouns have different suffixes for declension in the singular, and in the plural. A great number of masculine words in the singular are feminine in the plural. These words are called ambigens . Declension position is at the end of the word, except in the case of compound words where declension at the end of the first component (cilido, cilitdo; cilado, cilësdo; kushdo, kujtdo[4]).

Foreign Named Entities are transcribed according to Albanian phonetic: Meksika *Mexico,* Xhorxh Bernard Shou *George Bernard Shaw,* Penxhabi *the Punjab*, Pirenejtë *the Pyrenees*. Proper names are flexed with the same paradigms than other nouns. So do acronyms, but the flexion is preceded by a dash OKB-ja (nominative), OKB-në (accusative).

Adjectives have masculine, feminine, singular and plural forms. When an adjective without an article is after a noun, it is not declined; it is just flexed according to gender and number. It is the reverse if the adjective precedes the noun: the adjective itself is flexed and the noun is not.

- **Articulated Words : Some Nouns, Adjectives and Pronouns**

Most adjectives, some nouns and many pronouns are preceded by a particle called nyjë [5] and noted (i,e) in the dictionary. (i for masculine, e for feminine). Such words are called articulated words . These particles have declensions. These declensions vary according to the place of the articulated adjective or articulated noun in nominal syntagm. We have built inflectional graphs for articulated words.

---

[4] All three mean *anyone*. It is the same phenomenon than *auquel* and *auxquels* in French..

[5] The usual translation is article , but it is very different from an article in English or in French, so we use the word particle .

- **Inflectional Graphs**

We have studied the typology of singular and plural paradigms. With the exception of approximately thirty graphs, which require a specific graph each, there are about sixty graphs for masculine singular nouns and forty for feminine singular nouns. We have about eighty graphs that depict paradigms for the plural forms of the nouns. This number is increased by the problem of digraphs, as we explain below.

From description + entry, NooJ builds a list of declined forms. The result can be listed in a file (its extension is .flx) for verification. It is compiled and minimized in a Finite State Automata, for use during lexical analysis.

**Table 3.1.** Part of flex file for an ambigen word: agim,-i (-e, -et) *dawn*

| Entry | Flexions in the singular, define forms |
|---|---|
| agim,N+FLX=NS2_t+m+s | agimi,agim,N+FLX=NS2_t+m+s+emer+shquar<br>agimin,agim,N+FLX=NS2_t+m+s+kallez+shquar<br>së agimit,agim,N+FLX=NS2_t+m+s+gjin+shquar<br>të agimit,agim,N+FLX=NS2_t+m+s+gjin+shquar<br>i agimit,agim,N+FLX=NS2_t+m+s+gjin+shquar<br>e agimit,agim,N+FLX=NS2_t+m+s+gjin+shquar<br>agimit,agim,N+FLX=NS2_t+m+s+dhan+shquar<br>agimit,agim,N+FLX=NS2_t+m+s+rrjedh+shquar |
| | **Flexions in the plural, undefined forms** |
| agim,N+FLX=NPL7+f+p | agime,agim,N+FLX=NPL7+f+p+kallez+pashquar<br>agime,agim,N+FLX=NPL7+f+p+emer+pashquar<br>e agimeve,agim,N+FLX=NPL7+f+p+gjin+pashquar<br>i agimeve,agim,N+FLX=NPL7+f+p+gjin+pashquar<br>të agimeve,agim,N+FLX=NPL7+f+p+gjin+pashquar<br>së agimeve,agim,N+FLX=NPL7+f+p+gjin+pashquar<br>agimeve,agim,N+FLX=NPL7+f+p+dhan+pashquar<br>agimeve,agim,N+FLX=NPL7+f+p+rrjedh+pashquar<br>agimesh,agim,N+FLX=NPL7+f+p+rrjedh+geg+pashquar |

- **The Problem with Digraphs.**

Some rules like go left one letter , delete one letter , have to be redefined as go left two characters , delete two characters , according to the fact that one letter has two characters. And for the computer it is two letters! The whole set of words that obey the regular paradigm, has to be dispatched between the one character paradigm and the double character paradigm.

**Table 3.2.** Regular paradigms vs. single char. and double char. paradigm.

| paradigm | rule | example | translation |
|---|---|---|---|
| **paradigm 1** | Insert a, go left 1 letter, delete 1 letter | | |
| 1 char. paradigm 1 | $a<L><B>$ | motër → motra | *sister* |
| 2 char. paradigm 1 | $a<L2><B>$ | vjehërr → vjehrra | *mother in law* |
| **paradigm 2** | Insert i go left 1 letter, delete 1 letter | | |
| 1 char. paradigm 2 | $i<L><B>$ | gjarpër → gjarpri | *snake* |
| 2 char. paradigm 2 | $i<L2><B>$ | vjehërr → vjehrri | *faiher in law* |
| **paradigm 3** | delete 1 letter, insert j | | |
| 1 char. paradigm 3 | $<B>j$ | bir → bij | *son* |
| 2 char. paradigm 3 | $<B2>j$ | djall → djaj | *devil* |
| **paradigm 4** | Insert a, go left 1 letter, delete 1 letter, insert o | | |
| 1 char. paradigm 4 | $a<L><B>o$ | heq → hoqa | *To dig out* |
| 2 char. paradigm 4 | $a<L2><B>o$ | hedh → hodha | *To throw* |

Nouns ending by ër, or ërr, or ël, or ëll, construct their defined form in *loosing ë* and receiving a or i. See paradigms 1 and 2 in Table 3.2. Some nouns change their ending from the singular into the plural. For example the two words *bir* and *djall* drop their last letter and receive j. But the last consonant of *djall* is a double letter, so we need two paradigms for this transformation. See paradigm 3 in Table 3.2.

## 4. Building Dynamic Dictionary with Morphological graphs

A dictionary is a finite set of words; when there are paradigms that allow building infinite lists of words −called *open lists*-, they cannot be listed into dictionaries. These forms can be a form that is built from one or more words or affixes. They can be concatenated, or separated by a typographical element -an apostrophe, a dash or simply a space-and in some cases two different words are contracted, but this sometimes includes an apostrophe. We have to study the composition XY of two words X and Y in Albanian. We call them 'XY words'.

## 4.1 Numbers and XY Words with Numbers.

- **The cardinal numbers**

The cardinal numbers are partly concatenated e.g. 5: pesë, 10: dhjetë, 50: pesëdhjetë, 40: dyzet, '*hundred*': 'qind', 500: pesëqind. They can be compounds: 41: dyzet e një, 55: pesëdhjetë e pesë, 555: pesëqind e pesëdhjetë e pesë ("e" means and). NooJ can recognize cardinal numbers with *syntactical graphs*, as in French.

- **The ordinal numbers**

The ordinal numbers are articulated words fully concatenated e.g.:5[th]: (i,e) pestë, 41[th]: (i,e) dyzetenjëtë, 50[th]: (i,e) pesëdhjetë, 555[th]: (i,e). pesëqindepesëdhjetëepestë. The second component is recognized by a *morphological graph*. Fig. 4.1. the graph computes the number and adds it as a feature.

Roman numerals are recognized by the same *morphological graph* than in French.

**Fig. 4.1.** Morphological graph for ordinal numbers.

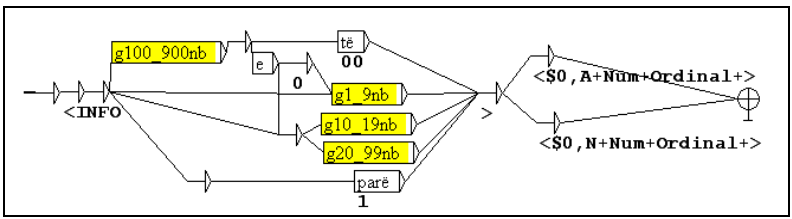

- **XY Words with Numbers**

There is a third form, that is like fully concatenated cardinal numbers. It is used as the first component X in XY words expressing a quantity of something: e.g. *fifty-five years old* is said pesëdhjetëepesëvjeçar . This is a single word, the first part of which is the concatenation of 55 pesëdhjetë e pesë followed by vjeçar that is not a word, but a component that means *aged of*. This construction is very productive. Any age can be expressed by a single word, as well as any number of floors, any number of months, of hours, of teeth etc. E.g.: *fifty floors*, pesëdhjetëkatësh, *twelve months*: Dymbëdhjetëmujor, *three teeth*: tredhëmbësh, *four motors*: katërmotorësh, two *times:* dyfish.

A lot of words have a  derived form  that can enter such a compound. The morphological graphs of NooJ can recognize dynamically such words as nouns, adjectives, adverbs, or verbs: it is possible to say not only *to quadruple* katërfishoj, but something like  t-uple  for any value of t.

Example of words recognized by the graph in Fig. 4.2
*dyfish,2fish,N*                              *dyfish,2fish,ADV*
*dyfishoj,2fishoj,V*                        *dymbëdhjetëmujor,12mujor,A+m*
*dyqindedyvjeçar,202vjeçar,A+m*   *katërfish,4fish,ADV*
*katërfish,4fish,N*                         *katërfishoj,4fishoj,V*

**Fig. 4.2.** Morphological graph for  XY words  where X is a number.

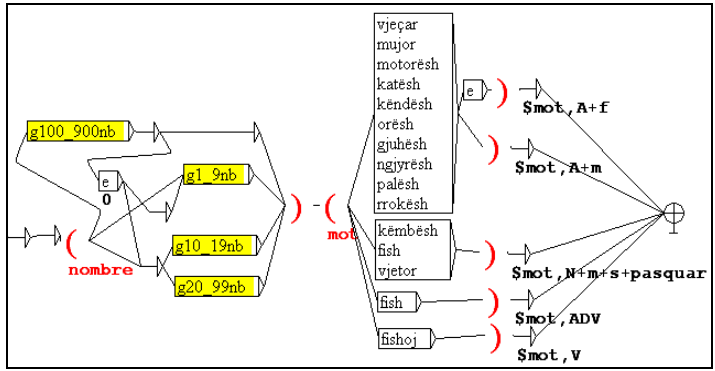

## 4.2  XY Words  Built with affixes.

Y can be a verb, a noun, or an adjective and X can be an affix, a prepositional form, a noun, an adjective or an adjective finished by an  o  e.g. leksiko-gramatikore *lexical-grammar*, italo-shqip *Italiano-Albanian*. About seventy affixes are listed in the Albanian dictionary. E.g.: gjysmë *half*, bashkë *together* bashkëbisedim *conversation*, pa *not*, aftë (i,e) *able* vs paaftë (i,e) *unable*, nën *under*, kuptoj *to understand*, nënkuptoj *to suggest*. This construction is very productive. An interesting property is that words are concatenated most often without contraction or modification. That makes them easy to recognize.

We have part the list of affixes between prefixes the biggest part-and the suffixes forms that are  fob : anglofob, gjermanofob, and  fobi : bakterofobi, fotofobi. We have already seen other suffixes in  XY words  with numbers.

A morphological graph can recognize several parts in a word and can compare them either to a specified list, or to a subset of words. We can insert parentheses with a name, like M1 and M2 and the word extracted in the loop with <L> is associated to the variable M2. Then M2 is compared to words in the dictionary. In case of success, the word is recognized and receives the lexical features $1L and the syntactic features $1S of M1. See Fig. 4.3.

**Fig. 4.3.** Graph for XY words where Y is a verb or a noun, and X is in a list.

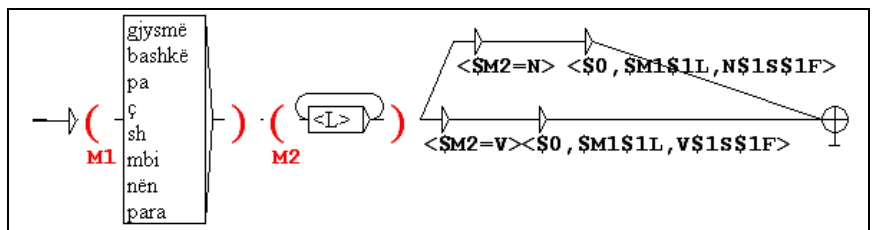

Example of wordsrecognized by this graph:
bashkëbisedimin,bashkëbisedim,N+m+s+kallez+shquar *conversation* *(*bashkë *together* bisedim *talk*)
mbijetonte,mbijetoj,V+Ind+3+s+I *to survive* (mbi *on* jetoj *to live*)
parashikoj,V+PR+Ind+1+s *to foresee* *(*para *before* shikoj *to see*)

## 4.3 Morphological Graphs For Imperative Tense with Clitics

As for the imperative form, the verb can concatenate with the clitic complement. In the plural the clitic can be inserted inside the imperative verb: trego *tell,* më trego or tregomë *tell me*, in the plural form the clitic is included inside the verb: *let you tell it* "e tregoni" → tregoj**e**ni. The letter j can be added in some forms. The figure 4.4. shows the results on examples.

**Fig. 4.4.** Morphological Graph to separate clitic and imperative verb.

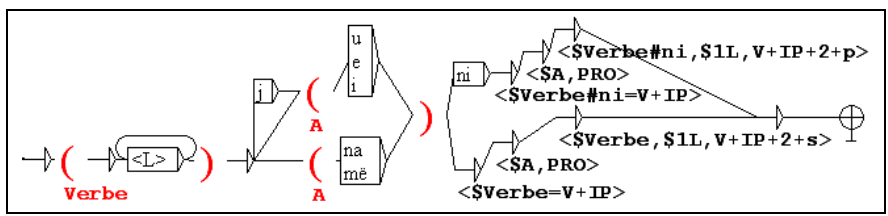

| tregojeni | merri! | hapini |
|---|---|---|
| 0   0,1 | 0   0,1 | 0   0,1 |
| e,PRQ  tregoj,V+IP+2+p | i,PRQ  marr,V+IP+2+s | i,PRQ  hap,V+IP+2+p |

## 5. X-Y Words Recognized by Syntactic Grammars

The syntactic graphs are used to recognize grammars involving several tokens. They can be organized into a hierarchy. So a syntactic grammar can use the result of an other one. This is very usual. We want to point out some cases very briefly.

### 5.1. Ordinal numbers

In figure 4.1. we show a graph that recognizes the second component of an ordinal number. For the whole ordinal number, that is a compound with a particle i or e, we use a syntactic graph.

### 5.2. X-X Words

Some words are built with a repetition of the same component. It is very common for onomatopoeias: e.g. tang-tang. But such words can be adverbs or adjectives. Nooj can point them out, and then we must validate the proposition. So we have introduced features such as + $hypo_n$,

**Fig. 5.1.** Syntactic grammar for X-X words .

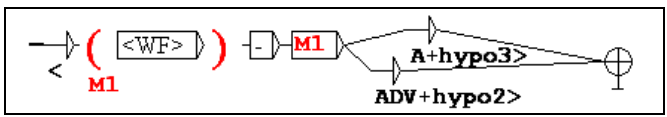

As in other languages, syntactic grammars can be used to recognize compound tenses.

### 5.3. Features in the Dictionaries

**Properties Definition**. The resulting dictionaries will be compiled as Finite State Transducers. They will register all words, plus their category and all their features, grammatical and linguistic. An important part of the work is to organize the features and tags. The aim is double: firstly, it will be necessary to display the annotations dictionary as a table; secondly this is necessary to be able to improve the dictionary. Whenever it is necessary to evolve syntactical and semantic tags, the dictionary will be modified. It is important to be able to adapt it easily. For example, a new study can make it necessary to define two tags instead of one. So the set of features and tags must be carefully registered and regularly updated.

**Table 5.**1. Properties definitions

| | |
|---|---|
| V_Pers = 1 + 2 + 3;<br>V_Nb = s + p;<br>V_Zgjedhimi = P + PP + PR + PS + I+ F;<br>V_Mënyra = Ind + Subj + Dëshirore + Habitore + IP + Kusht;<br>V_Trajta = NA + veprore + joveprore;<br>N_Nb = s + p;<br>N_Gender = m + f + as;<br>N_Shquar = shquar + pashquar;<br>N_Rasa = emer + rrjedh + gjin + kallez + dhan + rrjedh;<br>N_Ei = ei; | PREP_Rasa = emer + rrjedh + gjin + kallez<br>A_Nb = s + p;<br>A_Gender = m + f;<br>A_Rasa = = emer + rrjedh + gjin + kallez + dhan;<br>A_Ei = ei;<br>A_Shquar = shquar + pashquar;<br>DET_Nb = s + p;<br>DET_Genre = m + f;<br>PRO_Pers = 1 + 2 + 3;<br>PRO_Nb = s + p;<br>PRO_Rasa = emer + rrjedh + gjin + kallez + dhan;<br>PRO_Shquar = shquar + pashquar; |

We have said that our dictionary is only a syntactical one. We must precise that each time that we have the opportunity to acquire some words not only with syntactic features, but as well with semantic information, we do it. As a result we have some semantic tags. It is all the more important to register them in a properties definitions file. It will be necessary to add semantic information in a next step.

## Conclusion

We conclude that the NooJ s graphs -morphological graphs and syntactic grammars- are very efficient for the automatic processing of the Albanian language. We are able to identify  XY words  created by simple concatenation, as well as  X-Y words  built with an hyphen. We can recognize the words built with numbers. We can recognize dynamically different kind of open lists of words.

Our aim is to be able to automate this language but because some words and some forms in dialect are still being used, we have chosen to add some traditional forms (for example plural ablative Gheg in  sh , Gheg future and Gheg infinitive), that are commonly used. The choice of the features cannot be seen as achieved. Every new work can lead us to part set of words, to add new features or to transform them. The dictionary is growing bigger and bigger. But a big inconvenient is that texts older than 35 years will not be recognized, except partly.

It will be necessary to add entries from others Albanian dictionaries to this first electronic dictionary. We are mindful that there is still a lot to do.